\documentclass[letterpaper]{article} 
\usepackage{aaai23-r2hcai}  
\usepackage{times}  
\usepackage{helvet}  
\usepackage{courier}  
\usepackage[hyphens]{url}  
\usepackage{graphicx} 
\usepackage{makecell}
\usepackage{array}
\usepackage{caption} 
\usepackage{algorithm}
\usepackage{amsmath}
\usepackage{algorithmic}
\usepackage[utf8]{inputenc}
\usepackage[T5]{fontenc}
\usepackage{babel}
\usepackage{fancyhdr}
\usepackage{newfloat}
\usepackage{listings}

\usepackage{xcolor}
\usepackage{url}

\DeclareCaptionStyle{ruled}{labelfont=normalfont,labelsep=colon,strut=off} 
\floatstyle{ruled}
\newfloat{listing}{tb}{lst}{}
\floatname{listing}{Listing}
\usepackage{fancyhdr}
\fancypagestyle{firstpagehf}
{
   \fancyhf{}

   \fancyfoot[C]{\thepage}
    
}

\title{Enhancing OCR for Sino–Vietnamese Language Processing via Fine-Tuned PaddleOCRv5}

\author{
    Minh Hoang Nguyen$^{1, 2}$* \quad
    Su Nguyen Thiet$^{1, 2}$
}

\affiliations {
    \textsuperscript{\rm 1} Faculty of Information Technology, University of Science, Ho Chi Minh City, Vietnam \\
    \textsuperscript{\rm 2} Vietnam National University, Ho Chi Minh City, Vietnam \\
    \{24C15049, 24C15023\}@student.hcmus.edu.vn
}

\usepackage{bibentry}

\begin{document}
\thispagestyle{firstpagehf}
\maketitle

\begin{abstract}
Recognizing and processing Sino-Nom (Han–Nom) texts play a vital role in digitizing Vietnamese historical documents and enabling cross-lingual semantic research. However, existing OCR systems struggle with degraded scans, non-standard glyphs, and handwriting variations common in ancient sources. In this work, we propose a fine-tuning approach for PaddleOCRv5 to improve character recognition on Han–Nom texts. We retrain the text recognition module using a curated subset of ancient Sino-Nom manuscripts, supported by a full training pipeline covering preprocessing, LMDB conversion, evaluation, and visualization. Experimental results show a significant improvement over the base model — exact accuracy rises from 37.5\% to 50.0\%, particularly under noisy image conditions. Furthermore, we develop an interactive demo that visually compares pre- and post-fine-tuning recognition results, facilitating downstream applications such as Han–Vietnamese semantic alignment, machine translation, and historical linguistics research. The demo is available at: \url{https://huggingface.co/spaces/MinhDS/Fine-tuned-PaddleOCRv5}

\end{abstract}

\section{I. Introduction and Related Work}
In the context of digital preservation and cultural heritage research, the construction of electronic corpora for linguistic studies, machine translation, and the analysis of ancient texts has become increasingly crucial. In Vietnam, a particularly valuable source of historical material is the body of Sino-Nom documents, with many characters borrowed or adapted from ancient Chinese texts, exemplified by the monumental work \textit{Khâm Định Việt Sử Thông Giám Cương Mục}~\cite{sử2014kham}, compiled during the Nguyễn dynasty with both Han and Vietnamese (Quốc ngữ) versions.\footnote{\url{https://quangduc.com/a4672/kham-dinh-su-viet-thong-giam-cuong-muc-pdf}}. However, the automatic extraction of information from these historical sources remains technically challenging, particularly in the task of character recognition from degraded manuscript images.

\begin{figure}[ht]
    \centering
    \includegraphics[width=\linewidth]{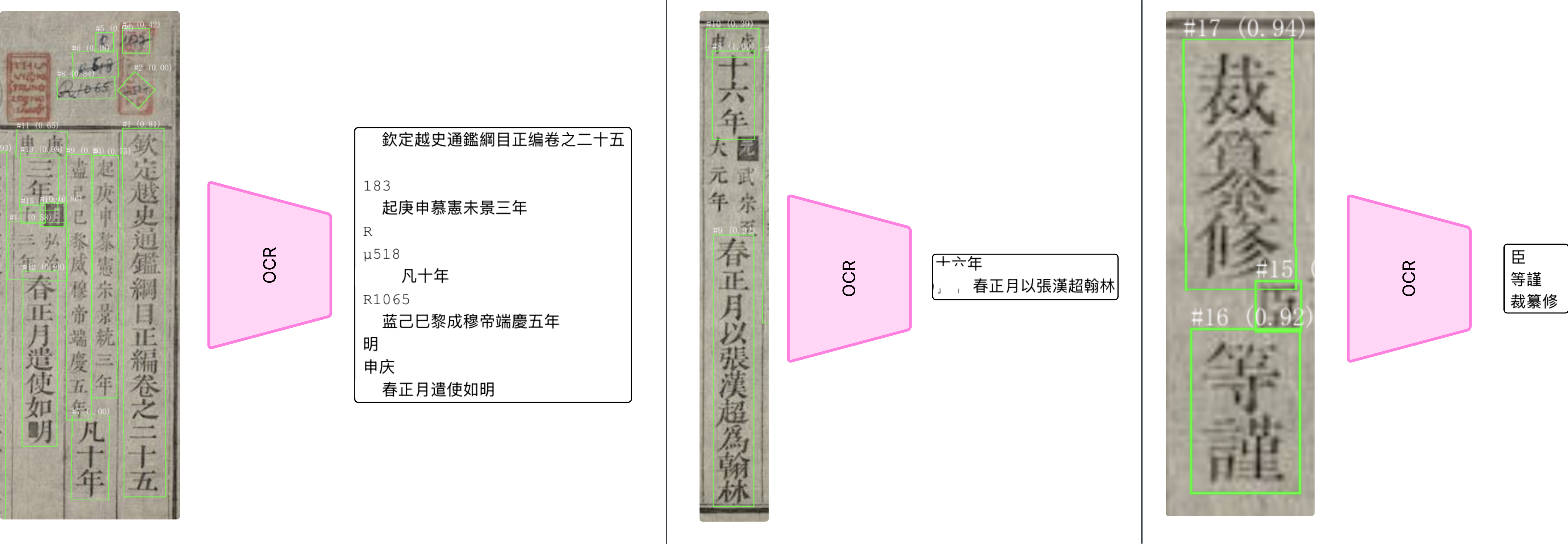}
    \caption{Examples of OCR errors in ancient Han–Nom texts: noisy characters, missing lines, and sentence merging.}
    \label{fig:ocr-error}
\end{figure}

\begin{figure}[ht]
    \centering
    \includegraphics[width=\linewidth]{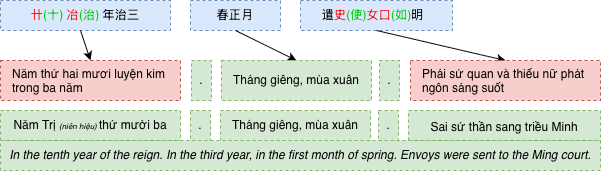}
    \caption{An example of an alignment error caused by incorrect character mapping due to OCR mistakes.}
    \label{fig:alignment-error}
\end{figure}

Research on the optical character recognition (OCR) of Han–Nom texts remains limited compared with modern languages. Zhang et al.~\cite{zhang2022chinese} proposed an OCR model for ancient Chinese using a CNN with an attention-based sequence decoder, but their dataset primarily contained clean printed texts that do not reflect the noise level of archived manuscripts. Nguyen et al.~\cite{nguyen2017tens} presented one of the first systems for Han–Nom character recognition using a CNN with rule-based post-processing, yet without fine-tuning on real-world historical data. Modern frameworks such as PaddleOCR~\cite{du2021ppocr, huang2022ppocrv3} offer strong recognition performance for contemporary Chinese, particularly in versions \texttt{PP-OCRv4/v5} with a two-stage detection–recognition pipeline. However, to our knowledge, no prior work has explored fine-tuning PaddleOCR specifically for ancient Vietnamese Han texts. In related areas, \texttt{Bertalign}~\cite{dou2022bertalign} has demonstrated promising results for Han–Vietnamese semantic alignment, yet its accuracy is directly constrained by OCR quality.

\begin{figure*}[!ht]
\centering
\includegraphics[width=7in]{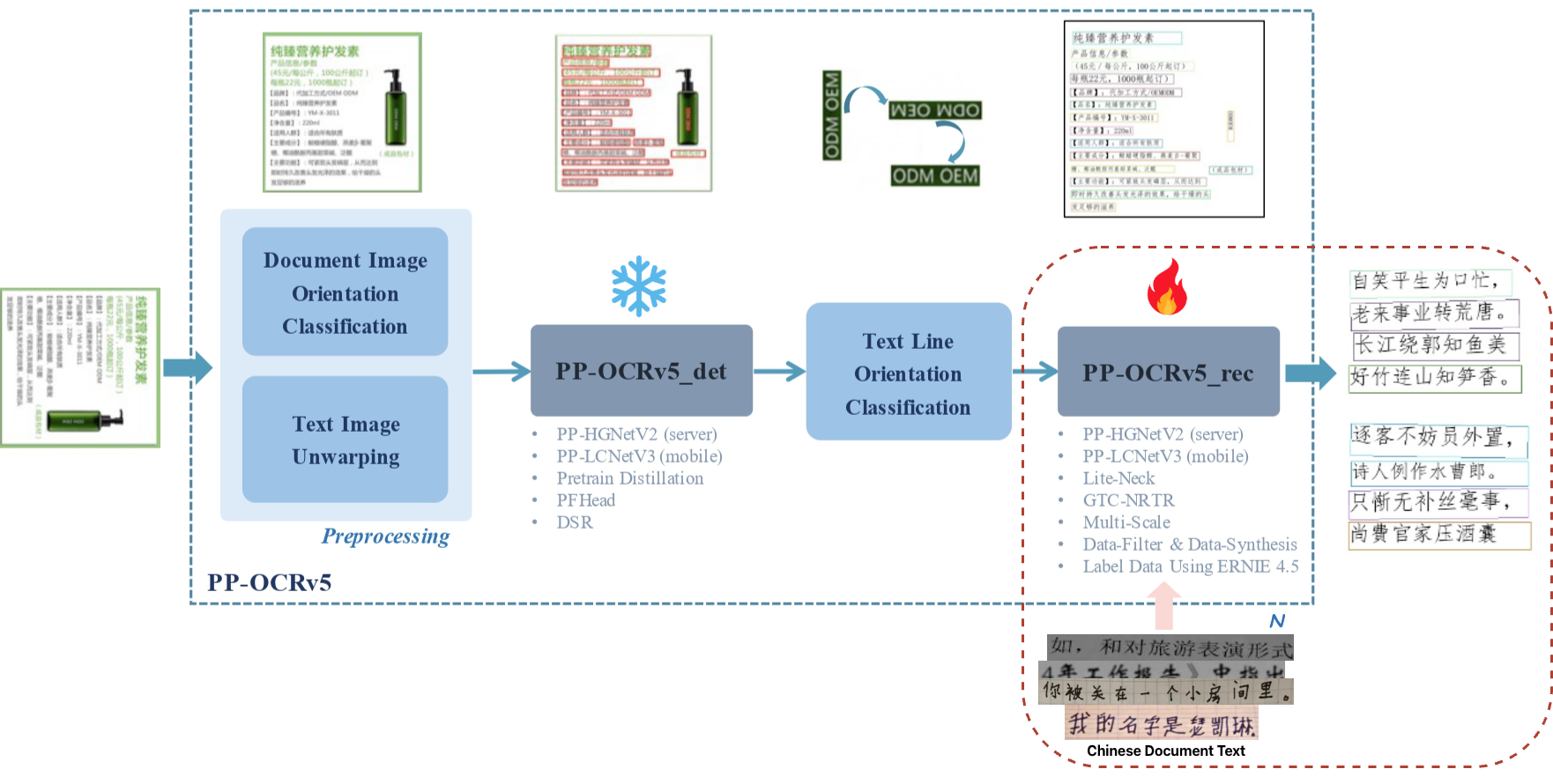}
\caption{Fine-tuning pipeline of PaddleOCRv5 for character recognition on ancient Han–Nom documents.}
\label{fig2}
\end{figure*}

Despite the high accuracy achieved by modern OCR systems such as \texttt{PaddleOCR} on standard Chinese benchmarks, their direct application to ancient documents still suffers from significant errors—including character noise, line loss, and incorrect sentence segmentation. These issues arise from the complex visual characteristics of historical sources: uneven scanning quality, ink stamps, narrow margins, mixed handwriting styles, and non-standard layouts. Such visual noise not only reduces recognition accuracy but also degrades the performance of downstream tasks like Han–Vietnamese alignment or semantic extraction.

Based on error analysis from our preliminary study (see Fig.~\ref{fig:ocr-error}, and Fig.~\ref{fig:alignment-error}), we observe that the default \texttt{PP-OCRv5} model is not optimized for ancient Han–Nom data. This paper therefore proposes a fine-tuning approach for PaddleOCRv5 using a curated dataset of handwritten Han–Nom texts. Our goal is to enhance recognition performance for Sino-Nom documents, enabling applications in knowledge extraction, translation studies, and cross-lingual semantic research. The main contributions of this paper are as follows:
\begin{itemize}
    \item We present a \textbf{fine-tuned OCR system specifically for Han–Nom (Sino–Vietnamese) texts}, leveraging a curated dataset of Vietnamese and broader Asian historical documents, addressing noise, degradation, and non-standard glyph variations common in ancient manuscripts.
    \item We propose both quantitative and visual evaluation methods to assess fine-tuning effectiveness—comparing pre- and post-training results on real historical images using metrics such as Character Accuracy and Confidence Score, supported by an demo.
    \item Experimental results confirm that our approach effectively improves OCR accuracy on Han–Nom datasets.
\end{itemize}

\section{II. Methodology}

\subsection{2.1 Fine-Tuning the Character Recognition Module of PP-OCRv5}

In this study, we focus on fine-tuning the recognition module of PP-OCRv5~\cite{du2021ppocr} (\texttt{PP-OCRv5\_rec}) to improve the recognition performance for Han–Nom (Sino–Vietnamese characters) texts (see Fig.~\ref{fig2}). The recognition module of PP-OCRv5 adopts a dual-branch architecture consisting of:
\begin{itemize}
    \item \textbf{GTC-NRTR}~\cite{hu2020gtc} — a sequence-to-sequence model with an attention mechanism that enhances sequence modeling capability during training.
    \item \textbf{SVTR-HGNet}~\cite{du2022svtr} — a lightweight branch built on the PP-HGNetV2 backbone, optimized with Connectionist Temporal Classification (CTC) loss~\cite{graves2006connectionist} for fast inference.
\end{itemize}

\begin{figure*}[!ht]
\centering
\includegraphics[width=7in]{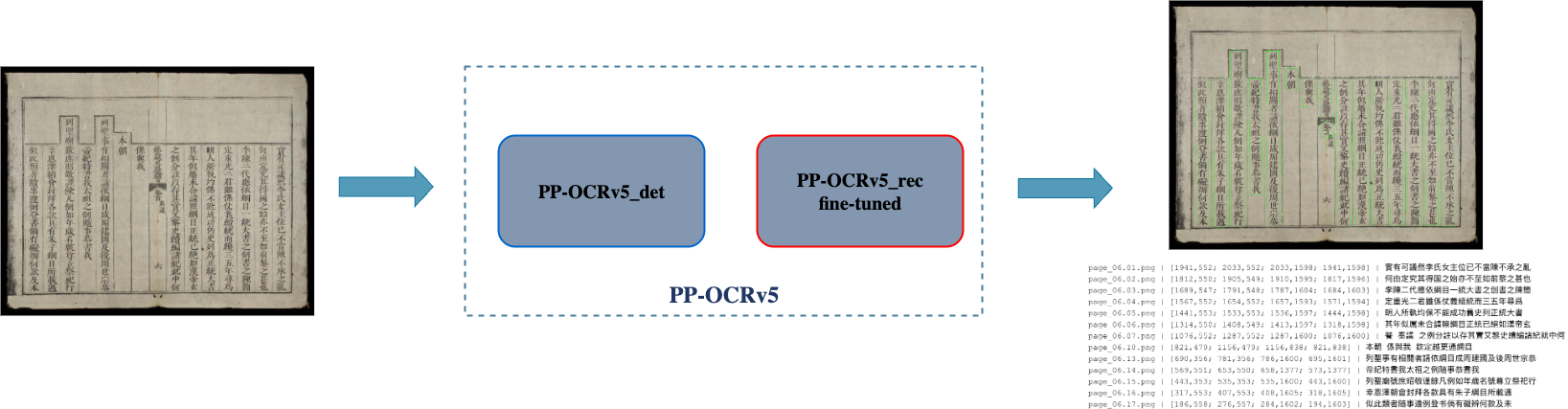}
\caption{Inference pipeline of the fine-tuned PaddleOCRv5 model for ancient Han–Nom documents.}
\label{fig:infer}
\end{figure*}

During training, the GTC-NRTR branch acts as a \textit{teacher} that guides the SVTR-HGNet branch (\textit{student}) through a knowledge distillation process defined as:

\begin{equation}
    \mathcal{L}_{\text{total}} = \lambda_1 \cdot \mathcal{L}_{\text{CTC}} + \lambda_2 \cdot \mathcal{L}_{\text{KD}}
\end{equation}

where \( \mathcal{L}_{\text{CTC}} \) is the CTC loss applied to the student branch, \( \mathcal{L}_{\text{KD}} \) represents the distillation loss between the two branches, and \( \lambda_1, \lambda_2 \) are hyperparameters balancing their contributions. At inference time (Fig.~\ref{fig:infer}), only the SVTR-HGNet branch is used, ensuring efficient runtime performance.

\subsection{2.2 Training Data and Preprocessing}

Constructing a large-scale annotated corpus from historical documents such as the \textit{Khâm Định Việt Sử Thông Giám Cương Mục} is resource-intensive. Therefore, to enable controlled fine-tuning experiments, we employed an existing publicly available handwritten dataset — a subset extracted from the \texttt{Chinese Document Text Recognition Dataset}\footnote{\url{https://github.com/YCG09/chinese_ocr}} — which contains approximately 3.64 million synthetic text-line images generated from both classical and modern Chinese corpora. These samples were produced through visual transformations such as font variation, resizing, blurring, geometric distortion, stretching, and random noise injection.

\begin{table}[ht]
\centering
\caption{Statistics of the dataset subset used for fine-tuning.}
\label{tab:data-stats}
\begin{tabular}{|l|r|}
\hline
\textbf{Component} & \textbf{Quantity} \\
\hline
Total images in original dataset & 3,640,000 \\
Images used for fine-tuning & 400,000 \\
Images used for evaluation & 100,000 \\
Average label length & 5.3 characters \\
Unique characters & 476 \\
Image resolution & $280 \times 32$ \\
Image format & RGB \\
\hline
\end{tabular}
\end{table}

Each image contains approximately ten consecutive characters cropped from arbitrary text lines. The full dataset includes around 5,990 unique characters, covering Chinese characters, Latin letters, digits, and punctuation marks. Images are standardized to a resolution of \(280 \times 32\) pixels, and labels follow the \texttt{image\_path<tab>groundtruth} format, compatible with PaddleOCR’s training interface.  

\begin{figure}[ht]
    \centering
    \includegraphics[width=\linewidth]{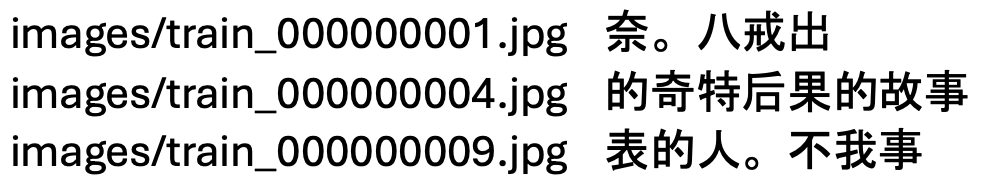}
    \caption{Examples of preprocessed Han–Nom text-line samples used for fine-tuning.}
    \label{fig:ocr-sample}
\end{figure}

Due to computational limitations, we randomly sampled 1,000,000 images for training and 100,000 images for evaluation. After cleaning, the subset had an average label length of 5.3 characters and contained 476 distinct characters in total. The pre-processing and data preparation pipeline includes:

\begin{enumerate}
    \item \textbf{Format conversion:} Extract image–text pairs and convert annotations to single-line \texttt{.txt} format (Fig.~\ref{fig:ocr-sample}).
    \item \textbf{Data cleaning:} Remove corrupted or unlabeled samples.
    \item \textbf{Character dictionary generation:} Build a character–index mapping table for CTC decoding.
    \item \textbf{Dataset splitting:} Randomly divide the cleaned data into training and validation subsets (10:1 ratio).  
\end{enumerate}

\subsection{2.3 Research Questions}

This study is guided by two key research questions:

\begin{itemize}
    \item \textbf{RQ1:} Does fine-tuning the recognition module on Han–Nom data improve OCR performance on real ancient texts?
    \item \textbf{RQ2:} How do visual and linguistic factors—such as noise, character rarity, and layout complexity—affect model performance?
\end{itemize}
To address these questions, we conduct both quantitative evaluation (Exact Accuracy, Partial Accuracy, and Average Confidence Score) and qualitative error analysis through visualization of recognition results, allowing us to identify the main causes of recognition failures.

\section{III. Experiments}
\subsection{3.1 Experimental Setup}

We fine-tuned the PP-HGNetV2 model\footnote{\url{https://github.com/PaddlePaddle/PaddleClas/blob/release/2.6/docs/en/models/PP-HGNetV2_en.md}} using a single NVIDIA A6000 GPU. The model was trained for 10 epochs with a batch size of 8. Input images were resized to a fixed resolution of \(280 \times 32\) and processed as RGB tensors with shape \([3, 32, 280]\) (channel–height–width order). 

Optimization was performed using the AdamW algorithm \cite{loshchilov2017decoupled} with an initial learning rate of 0.001. The training objective was the Connectionist Temporal Classification (CTC) loss \cite{graves2006connectionist}, which is well-suited for unaligned sequence recognition tasks. All other hyperparameters followed the default configuration of PaddleOCR \cite{du2021ppocr}.

\subsection{3.2 Main Results}

Table~\ref{tab:training-results} summarizes the model’s performance before and after fine-tuning. The fine-tuned model achieves consistent improvements across all evaluation metrics, including exact and partial accuracy, as well as confidence scores.

\begin{table}[ht]
\centering
\small
\caption{Performance results before and after fine-tuning.}
\label{tab:training-results}
\begin{tabular}{|>{\raggedright\arraybackslash}p{3.9cm}|c|c|}
\hline
\textbf{Metric} & \makecell[c]{\textbf{Before} \\ \textbf{Fine-tuning}} & \makecell[c]{\textbf{After} \\ \textbf{Fine-tuning}} \\
\hline
Exact Accuracy & 37.5\% & 50.0\% ↑ \\
Partial Accuracy & 58.2\% & 70.3\% ↑ \\
Average Confidence & 81.3\% & 91.1\% ↑ \\
Total Recognized Characters & 4,865 & 5,311 ↑ \\
\hline
\end{tabular}
\end{table}

\textit{Exact accuracy} denotes the proportion of samples whose predicted text exactly matches the ground truth. \textit{Partial accuracy} allows for minor deviations (e.g., one missing or substituted character). \textit{Average confidence} measures the model’s mean output probability for predicted sequences.

\begin{figure}[ht]
    \centering
    \includegraphics[width=\linewidth]{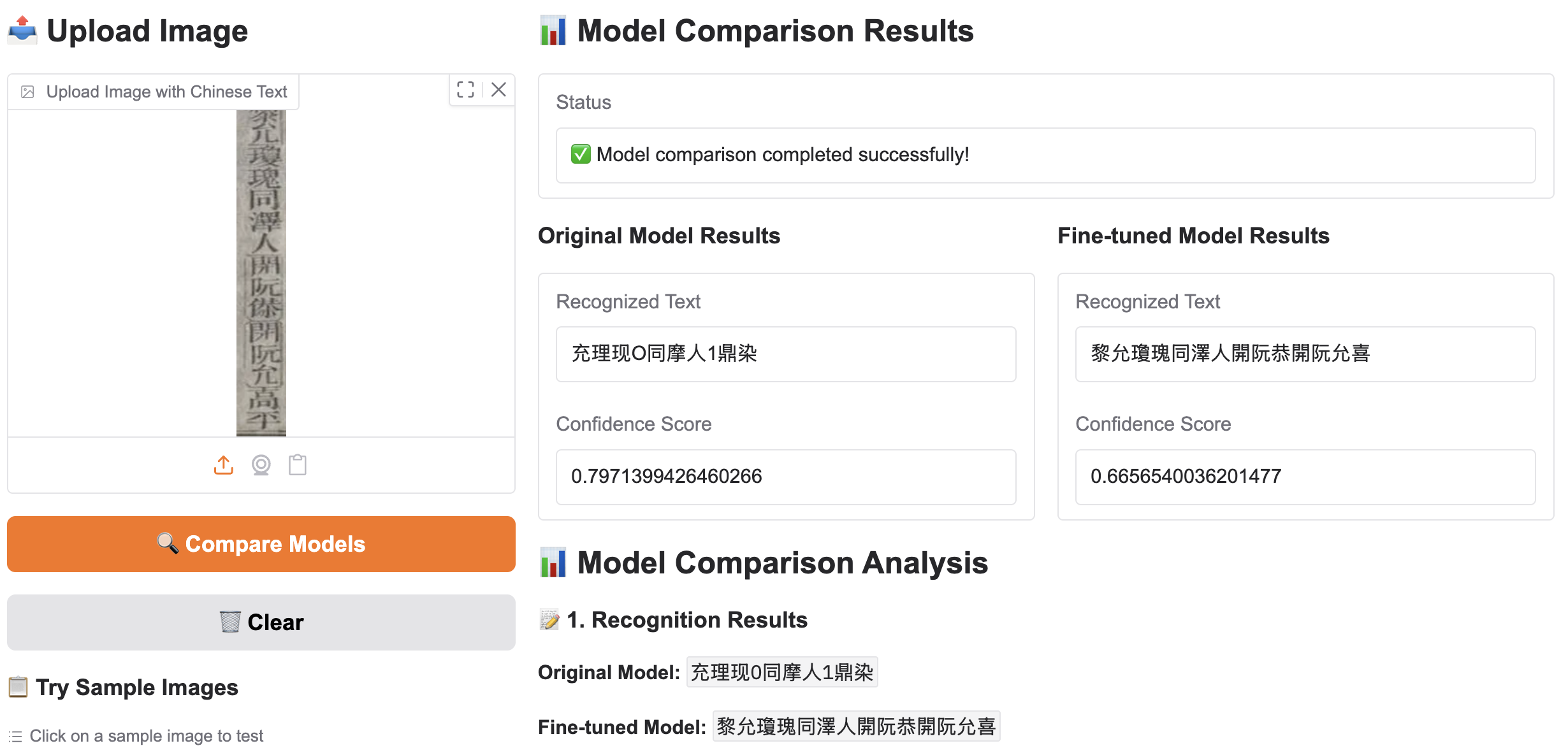}
    \caption{OCR Recognition examples before fine-tuning (left) and after fine-tuning (right).}
    \label{fig:compare}
\end{figure}

As illustrated in Figure~\ref{fig:compare}, fine-tuning not only improves overall recognition accuracy but also enhances robustness on challenging samples of ancient Chinese texts—particularly those with handwritten characters, blurred strokes, or noisy backgrounds. Notably, the average confidence score increases from 81.3\% to 91.1\%, suggesting that the model becomes more certain in its predictions.

\subsubsection{Analysis by Research Questions}
\paragraph{RQ1.} \textit{Does fine-tuning improve character recognition on ancient Chinese (Sino-Nom) texts?}  
Yes. Fine-tuning yields a substantial gain in recognition performance, with exact accuracy increasing from 37.5\% to 50.0\%. This confirms that adapting the OCR model to domain-specific data (ancient Chinese texts) effectively enhances accuracy, even under limited data and suboptimal imaging conditions.

\paragraph{RQ2.} \textit{What factors most affect model performance?}  
Error analysis reveals three main sources of degradation:
\begin{itemize}
    \item \textbf{Image noise:} Blurred or stripe-degraded regions often lead to character substitution errors.
    \item \textbf{Rare characters:} Low-frequency ancient Chinese characters are difficult for the model to represent reliably.
    \item \textbf{Complex layout:} Misaligned or densely packed text lines cause segmentation errors during preprocessing.
\end{itemize}

\section{IV. System Implementation}

To demonstrate the practical effectiveness of the fine-tuned model, we developed an interactive demo system deployed on Hugging Face Spaces\footnote{\url{https://huggingface.co/spaces}}. The deployment uses a basic CPU environment (2 vCPUs, 16 GB RAM) and allows users to upload images of Sino-Nom text to obtain recognition results in real time.

\begin{figure}[ht]
    \centering
    \includegraphics[width=\linewidth]{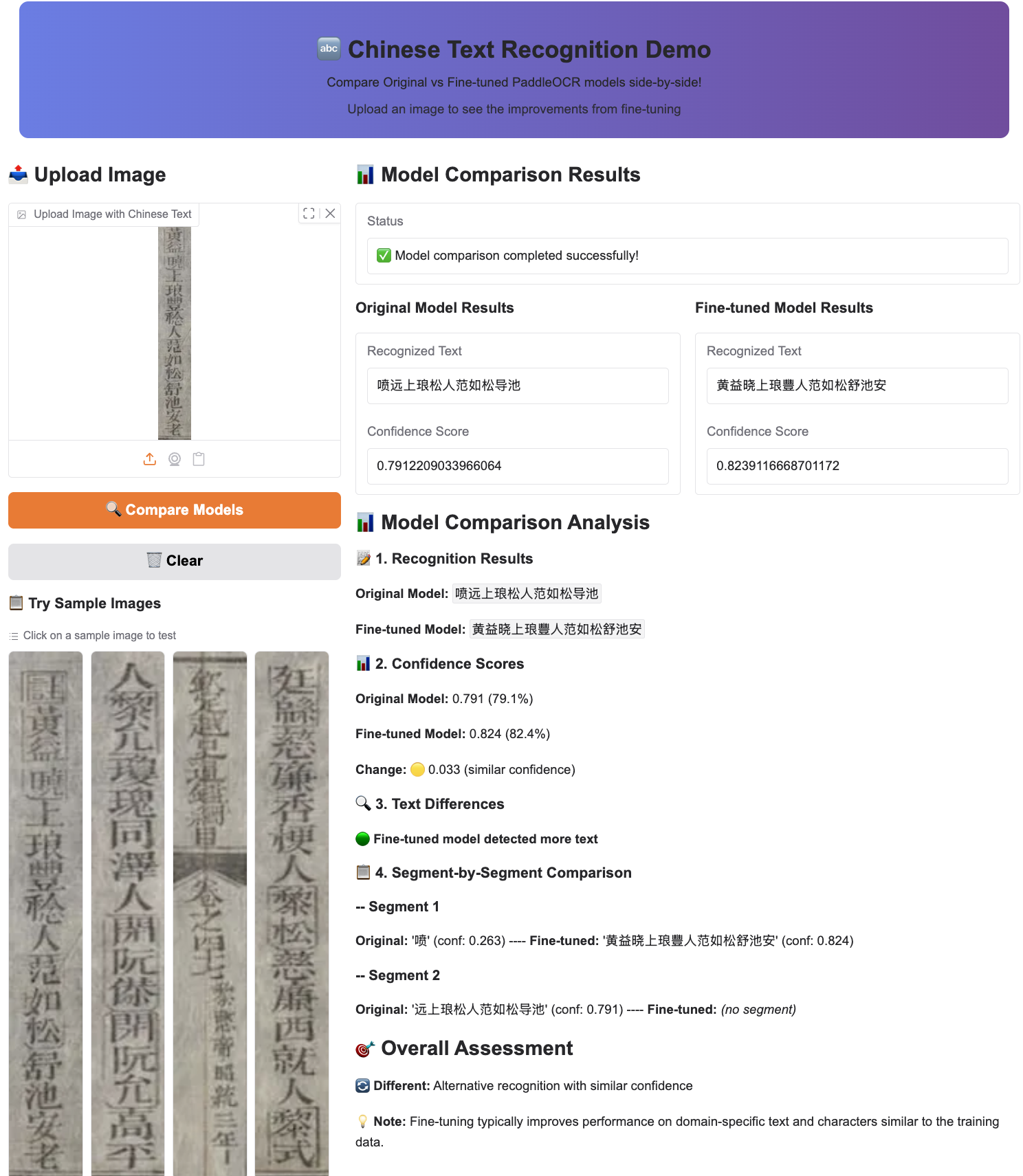}
    \caption{Demo interface for recognizing ancient Sino-Nom text using the fine-tuned model.}
    \label{fig:demo-ui}
\end{figure}

\subsection{4.1 System Architecture}
The demo system consists of two main components:
\begin{itemize}
    \item \textbf{Frontend (User Interface):} A lightweight web interface built with the Gradio library\footnote{\url{https://github.com/gradio-app/gradio}}, which enables users to upload images and visualize recognition results (before and after fine-tuning) directly on the image.
    \item \textbf{Backend (Inference Server):} A processing function that handles image preprocessing, text recognition via PaddleOCR, and response generation with recognized text sequences.
\end{itemize}

In testing, the average inference time ranged from 7 to 15 seconds per image.

\subsection{4.2 Core Functionalities}
\begin{itemize}
    \item Comparison between the baseline and fine-tuned models on the same input image.
    \item Display of recognition results with confidence scores, allowing users to assess prediction reliability.
    \item Support for image zooming, scrolling, and uploading of arbitrary images.
\end{itemize}

The demo system is publicly available at: \url{https://huggingface.co/spaces/MinhDS/Fine-tuned-PaddleOCRv5}.

\section{V. Conclusion}

This work presented a fine-tuning approach for the PaddleOCRv5 model to enhance character recognition on Han–Nom (ancient Sino–Vietnamese) texts, particularly those with low-quality images and irregular layouts. Experimental results demonstrated a substantial performance gain: the fine-tuned model achieved 50\% exact accuracy—an improvement of 12.5 percentage points over the baseline—while the average confidence score increased from 81.3\% to 91.1\%. We also developed an interactive demo platform to facilitate real-world evaluation, offering a practical tool for researchers working on tasks such as Han–Vietnamese machine translation, historical knowledge extraction, and cross-lingual semantic alignment. Future work will focus on several directions: \textbf{(1) Expanding the training corpus:} collecting and annotating additional real-world Han–Nom documents, including handwritten and printed materials; \textbf{(2) Two-stage fine-tuning (Detection + Recognition):} improving not only character recognition but also text-region detection for historical documents; \textbf{(3) Integrating semantic extraction pipelines:} combining OCR output with Han–Vietnamese translation and semantic alignment modules for end-to-end processing of ancient texts; and \textbf{(4) Cross-model evaluation:} comparing PaddleOCR with alternative OCR frameworks or self-supervised architectures to identify the most suitable model for ancient character recognition.

\subsubsection{Authorship contribution statement.} \textbf{Minh Hoang Nguyen} was Project Administration; Conceptualization; Writing – Original Draft; System Implementation; Methodology; Review \& Editing. \textbf{Su Nguyen Thiet} contributed to Methodology; Data Curation; Writing – Introduction \& Related Work; Experimental Summary; Writing – Review \& Editing.

\begin{figure*}[!ht]
\centering
\includegraphics[width=7in]{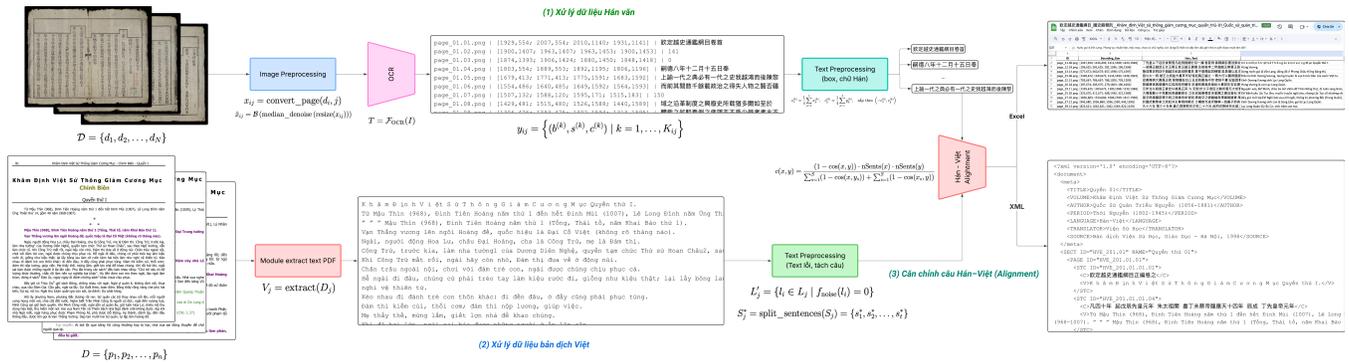}
\caption{Construction of Building the bilingual Han–Vietnamese pipeline.}
\label{fig:overview-alignment-pipe}
\end{figure*}

\bibliography{refs}
\bibliographystyle{ieeetr}

\newpage

\appendix
\section{Appendix A: Alignment Pipeline and OCR Fine-Tuning Impact}

In this appendix, we provide a detailed description of the Han–Vietnamese sentence alignment process and demonstrate how fine-tuning the PaddleOCRv5 model improves alignment quality. This section complements the main paper by highlighting technical details of preprocessing, alignment, and empirical results. For our experiments, we use the exemplary historical work \textit{Khâm Định Việt Sử Thông Giám Cương Mục}~\cite{sử2014kham}, based on PDF scans of Sino-Nom texts:
\begin{itemize}
    \item \textbf{Original Han text}: Stored at the National Library of Vietnam, in PDF image format, carved in 1884.
    \item \textbf{Vietnamese translation}\footnote{\url{https://quangduc.com/a4672/kham-dinh-su-viet-thong-giam-cuong-muc-pdf}}: Provided by the Vietnam Institute of Historical Studies. Official translation, Education Publishing House, Hanoi, 1998.
\end{itemize}

\section{A.1 Alignment Pipeline}

The construction of the bilingual Han–Vietnamese corpus follows three main stages (see Figure~\ref{fig:overview-alignment-pipe}):

\begin{enumerate}
    \item \textbf{Han text processing:} PDF scans of 51 historical volumes are converted into page images, preprocessed (resizing, denoising, binarization), and processed with OCR. Post-processing involves box correction, line merging, and sequence ordering according to traditional vertical writing (right-to-left, top-to-bottom). The result is a cleaned sequence of Han sentences $\mathcal{H} = \{h_1, h_2, \dots, h_N\}$.
    
    \item \textbf{Vietnamese translation processing:} PDF translations are extracted and cleaned to remove noise (titles, footnotes, page numbers) and split into sentences $\mathcal{V} = \{v_1, v_2, \dots, v_M\}$, ensuring consistent structure suitable for alignment.
    
    \item \textbf{Sentence alignment:} We apply \textbf{Bertalign} \cite{dou2022bertalign} to map Han sentences to Vietnamese sentences. Bertalign computes semantic embeddings for each sentence and uses dynamic programming to find the optimal 1–1, 1–n, or n–1 alignment:
    \[
    \text{alignment}(\mathcal{H}, \mathcal{V}) = \arg \min \sum c(x,y)
    \]
    where $c(x,y)$ is the cost function based on cosine similarity between embedding vectors, adjusted for sentence count in multi-sentence segments. The output is a set of aligned sentence pairs $\mathcal{A} = \{(h_i, v_j) \}$ with confidence scores $s_{ij} \in [0,1]$.
\end{enumerate}

\section{A.2 Integration of OCR Fine-Tuning}

OCR errors in Han text significantly affect alignment quality. Misrecognized characters can lead to incorrect semantic embeddings, resulting in lower alignment confidence or mismatched sentence pairs. To address this, we fine-tuned the character recognition module of PaddleOCRv5(see Section II in main paper).

After fine-tuning, the OCR output exhibits higher character accuracy and confidence, which directly improves downstream alignment. Specifically:

\begin{itemize}
    \item \textbf{Error reduction:} Character misrecognition in low-quality scanned pages decreases from 62.5\% to 50\%.
    \item \textbf{Alignment robustness:} More sentences are correctly matched to their Vietnamese counterparts, particularly in cases involving rare or archaic characters.
\end{itemize}

\section{A.3 Experimental Results}

We conducted experiments comparing alignment performance using OCR outputs from the \textbf{baseline} and \textbf{fine-tuned} models. Table~\ref{tab:alignment-results} summarizes the findings. Metrics include:

\begin{itemize}
    \item \textbf{Aligned sentence ratio:} Proportion of Han sentences successfully mapped to Vietnamese.
    \item \textbf{Average confidence:} Mean Bertalign alignment score across all sentence pairs.
    \item \textbf{Exact 1–1 alignment rate:} Fraction of sentences aligned in a strict 1–1 manner.
\end{itemize}

\begin{table}[ht]
\centering
\small
\caption{Effect of OCR fine-tuning on Han–Vietnamese sentence alignment.}
\label{tab:alignment-results}
\begin{tabular}{|>{\raggedright\arraybackslash}p{3.9cm}|c|c|}
\hline
\textbf{Metric} & \makecell[c]{\textbf{Baseline} \\ \textbf{OCR}} & \makecell[c]{\textbf{Fine-Tuned} \\ \textbf{OCR}} \\
\hline
Aligned sentence ratio & 61.8\% & 67.6\% ↑ \\
Average alignment confidence & 0.783 & 0.842 ↑ \\
Exact 1–1 alignment rate & 46.2\% & 50.4\% ↑ \\
Total aligned pairs & 15,987 & 17,455 ↑ \\
\hline
\end{tabular}
\end{table}

\noindent The results indicate a clear improvement in both the quantity and quality of aligned sentence pairs after fine-tuning OCR. Notably:

\begin{itemize}
    \item More sentences with rare or complex characters are correctly aligned.
    \item Average semantic similarity scores increase, reflecting higher alignment confidence.
    \item Downstream applications, such as machine translation or bilingual corpus construction, benefit from more reliable sentence pairs.
\end{itemize}

Fine-tuning the OCR model on Chinese Document Text Recognition Dataset substantially reduces recognition errors, which translates into improved semantic embeddings for alignment. Consequently, the Bertalign-based pipeline yields a larger number of high-confidence aligned sentences. This appendix demonstrates the direct impact of OCR enhancement on bilingual corpus quality, confirming that preprocessing and OCR fine-tuning are crucial steps in building reliable Han–Vietnamese datasets.

\end{document}